\documentclass[11pt, a4paper, gr]{google}
\usepackage[authoryear, sort&compress, round]{natbib}
\usepackage{microtype}
\usepackage{hyperref}
\usepackage{url}
\usepackage{booktabs}

\usepackage{lineno}

\usepackage{graphicx}

\usepackage{amsmath}
\usepackage{cleveref}

\usepackage{subcaption}
\usepackage{enumitem} % Required in preamble

\usepackage{tabularx}
\usepackage{booktabs}
\usepackage{xcolor}
\usepackage{wrapfig}
\usepackage[table]{xcolor} % Required for \rowcolors

\usepackage{listings}
\usepackage{caption}

\usepackage[T1]{fontenc} % Uses an encoding that supports straight quotes
\usepackage{textcomp}    % Provides additional text symbols
\usepackage{upquote}     % Forces straight quotes (" and ') in code blocks

\usepackage{listings}
\uselogo{} 

\title{Thinking to Recall: How Reasoning Unlocks Parametric Knowledge in LLMs}

\correspondingauthor{zorik@google.com, jherzig@google.com}

\reportnumber{0001} % Leave blank if n/a

\renewcommand{\monthyeardate}{}
\renewcommand{\today}{}

\author[1,2]{Zorik Gekhman}
\author[1]{Roee Aharoni}
\author[1]{Eran Ofek}
\author[3]{Mor Geva}
\author[2]{Roi Reichart}
\author[1]{Jonathan Herzig}

\affil[1]{\thepa{}{}}
\affil[2]{Technion - Israel Institute of Technology}
\affil[3]{Tel Aviv University}

\begin{abstract}
While reasoning in LLMs plays a natural role in math, code generation, and multi-hop factual questions, its effect on simple, single-hop factual questions remains unclear. Such questions do not require step-by-step logical decomposition, making the utility of reasoning highly counterintuitive. Nevertheless, we find that enabling reasoning substantially expands the capability boundary of the model's parametric knowledge recall, unlocking correct answers that are otherwise effectively unreachable. \textit{Why does reasoning aid parametric knowledge recall when there are no complex reasoning steps to be done?} To answer this, we design a series of hypothesis-driven controlled experiments, and identify two key driving mechanisms: (1) a computational buffer effect, where the model uses the generated reasoning tokens to perform latent computation independent of their semantic content; and (2) factual priming, where generating topically related facts acts as a semantic bridge that facilitates correct answer retrieval. Importantly, this latter generative self-retrieval mechanism carries inherent risks: we demonstrate that hallucinating intermediate facts during reasoning increases the likelihood of hallucinations in the final answer. Finally, we show that our insights can be harnessed to directly improve model accuracy by prioritizing reasoning trajectories that contain hallucination-free factual statements.\looseness=-1
\end{abstract}

\begin{document}

\maketitle

\section{Introduction}

Reasoning Large Language Models (R-LLMs) are trained to generate a long Chain-of-Thought with a step-by-step solution before predicting the final response \citep{guo2025deepseek,comanici2025gemini,openai2025introducing,yang2025qwen3}. While these models significantly improved performance on complex tasks such as math and coding, their impact on simple, single-hop factual questions is less intuitive since such questions do not require logical decomposition or complex multi-step reasoning. Nevertheless, we find that enabling reasoning substantially expands the model's parametric knowledge boundary, unlocking correct answers that are otherwise effectively unreachable.

\begin{figure}[t] % You can now use [h!] if you want it 'here'
    \centering
    \includegraphics[width=0.9\columnwidth, trim={0 0cm 0 0}, clip]{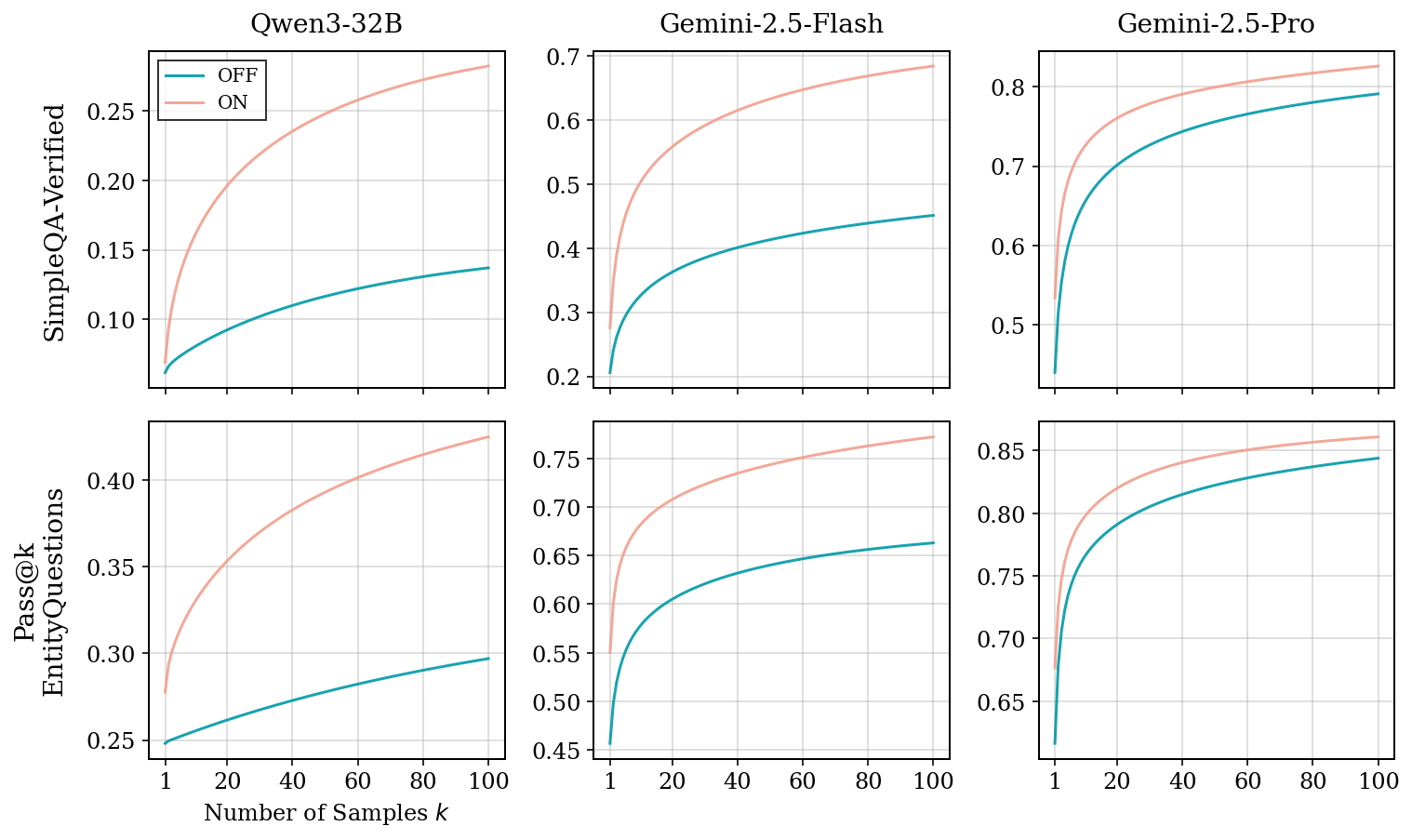}
% trim={<left> <bottom> <right> <top>}
     % Adjust this value until it looks right
    \caption{
    % Comparison of pass@$k$ scores for with reasoning \texttt{OFF} vs \texttt{ON}.
    Pass@$k$ curves across two closed-book QA benchmarks and three LLMs, comparing the same models with reasoning \texttt{OFF} vs \texttt{ON}.}
     % Adjust this value until it looks right
    \label{fig:on_vs_off}
    
\end{figure}

To study this effect, we use hybrid models where reasoning can be toggled \texttt{ON}/\texttt{OFF}, allowing to isolate its effect while controlling for the model's parametric knowledge. We probe the model's parametric recall capability boundary using the pass@$k$ metric \citep{yue2025does}, which estimates how coverage (the fraction of questions correctly answered by any sample) scales with the number of sampled answers. While both modes follow the expected scaling trend, the reasoning \texttt{ON} mode consistently outperforms reasoning \texttt{OFF} (\Cref{fig:on_vs_off}). Importantly, we find that these gains stem from improved knowledge recall rather than multi-hop reasoning, as questions labeled as ``multi-hop'' or ``requires reasoning'' in \texttt{SimpleQA-Verified} \citep{haas2025simpleqa} did not benefit from reasoning more than simple, single-hop questions. This raises the question of what mechanisms make reasoning helpful for parametric knowledge recall. Understanding them is critical, as it can inform improved training recipes, such as targeted process rewards, as well as more effective inference strategies.\looseness=-1

Through hypothesis-driven controlled experiments, we identify two key mechanisms. The first is a content-independent \textit{computational buffer} effect: R-LLMs can use the generated reasoning tokens to perform latent computation and refine their predictions. This is evidenced by their ability to benefit even from filler reasoning traces without semantic content. However, this effect plateaus below full reasoning performance, indicating that it alone cannot account for the observed gains. This motivates a closer examination of the semantic content of reasoning traces: since they do not contain step-by-step decomposition, it is unclear how their content could improve predictions beyond the computational buffer effect. In a carefully designed controlled experiment, we demonstrate the existence of a second, content-dependent \textit{factual priming} mechanism: The model engages in generative self-retrieval, constructing contextual bridges to the answer by recalling related facts. Interestingly, extracting a \textit{short} list of facts from the reasoning trace and rerunning the model with reasoning disabled conditioned on this list as additional context, recovers most of the pass@k gains of reasoning, providing strong evidence that the retrieved intermediate facts themselves are useful for recalling the correct answer.

However, the factual priming mechanism introduces a fundamental risk, since the facts are generated by the model itself and may therefore be hallucinated. We assess this risk using a large-scale auditing pipeline that verifies every intermediate fact in every sampled trajectory for each question, where each fact is evaluated via a dedicated, search-enabled verification call to Gemini-2.5-Flash \citep{comanici2025gemini}. This effort reveals a clear pattern: reasoning traces with hallucinated intermediate facts are substantially more likely to yield hallucinated final answers. Thus, generative self-retrieval is a powerful but fragile mechanism: it can surface latent knowledge, yet it also permits reasoning-stage errors to shape the final answer.

Finally, we demonstrate that these insights can be leveraged at inference time to directly improve model accuracy. By simulating a test-time selection strategy that prioritizes reasoning trajectories with hallucination-free factual statements, we highlight a promising direction for improving the factual reliability of R-LLMs.

We summarize our contributions as follows:

\begin{itemize}[itemsep=0.3ex, leftmargin=0.8cm, topsep=0.1ex]

    \item \textbf{Parametric Knowledge Boundary Expansion.} We show that reasoning substantially expands the model’s parametric recall boundary, evidenced by large pass@k gains.\looseness=-1\par

    \item \textbf{Question Complexity vs. Retrieval Difficulty.} We show that question complexity poorly predicts reasoning benefit, suggesting that reasoning gains are mainly driven by better parametric recall and not complex question decomposition.

    \item \textbf{Content-Independent Computation.} We show that R-LLMs use generated reasoning tokens as a computational buffer, performing latent computation independent of the reasoning trace’s semantic content.

    \item \textbf{Disentangling Content Effects.} We isolate the effect of the semantic content in reasoning traces and validate the factual priming hypothesis, showing that recalling related facts improves correct answer recall and drives the majority of reasoning gains.

    \item \textbf{Large-Scale Hallucination Audit.} We demonstrate that hallucinated facts in the reasoning trace increase the likelihood of final answer hallucinations.

    \item \textbf{Practical Implications.} We show that our insights can be operationalized: prioritizing reasoning  trajectories with desirable traits can yield substantial accuracy gains.

\end{itemize}

\vspace{-4pt}
\section{Setup}
\vspace{-4pt}
\label{sec:setup}

\textbf{Models.} We uses hybrid models where reasoning can be toggled \texttt{ON}/\texttt{OFF}, which are trained to recognize control tokens or system instructions dictating the inference mode: \texttt{ON} triggers the generation of a reasoning trace prior to the final response, whereas \texttt{OFF} suppresses this behavior. Using hybrid models, we isolate the effect of reasoning while controlling for the model's parametric knowledge. We use \textbf{Gemini-2.5-Flash}, \textbf{Gemini-2.5-Pro} \citep{comanici2025gemini}, and \textbf{Qwen3-32B} \citep{qwen3technicalreport}. See \S \ref{appendix:hyperparams} for more details and inference parameters.

\textbf{Datasets.} We use \textbf{SimpleQA} \citep{wei2024measuring} and \texttt{EntityQuestions} \citep{Entity_Questions}, two challenging closed-book QA datasets that allow us to balance realistic evaluation with experimental control; while \texttt{SimpleQA} contains realistic questions, \texttt{EntityQuestions} is based on question templates, which better decouples the difficulty of the question phrasing from the difficulty of parametric knowledge recall. For \texttt{SimpleQA}, we utilize \texttt{SimpleQA-Verified} \citep{haas2025simpleqa}, a subset of 1,000 examples filtered and corrected for increased reliability. For \texttt{EntityQuestions}, we follow \citet{gekhman2025insideout} and focus on four relations that have a large answer space and contain unambiguous answers, randomly sampling 250 examples from each to a total of 1,000 (see \S \ref{appendix:eq_rels} for details and \Cref{tab:relations} for the list of relations).

\textbf{Metric.} We use \textbf{pass@}$\mathbf{k}$ and motivate it in \S \ref{sec:capability_boundady}. It estimates the probability that at least one of $k$ sampled answers is correct. To compute it we must grade predicted answers' correctness by comparing to the ground truth, we do this using Gemini-2.5-Flash as autorater using the prompt from \citet{wei2024measuring} for \texttt{EntityQuestions} and its adapted version from \citet{haas2025simpleqa} for \texttt{SimpleQA-Verified}. 
 We use the unbiased estimation method from  \citet{chen2021evaluating}.

\section{Reasoning Expands The Model's Parametric Knowledge Boundary}
\label{sec:capability_boundady}
 % Adjust this value 

In this section we evaluate the potential of reasoning to improve parametric recall. We focus on \textit{capability boundary} and ask: Does reasoning enable the discovery of correct answers that are effectively unreachable without it, or does it mainly improve accuracy via better \textit{sampling efficiency}, by boosting the probability of (correct) answers that are already relatively likely under the model's policy? This distinction is important since if reasoning elevates the probability of correct answers with very low initial likelihood, their generation could then be further encouraged through training with verifiable rewards \citep{lambert2024tulu,guo2025deepseek}, or prioritized during deployment via inference-time techniques \citep{snell2025iclr, wu2025iclr, brown2024large}. We use the \textbf{pass@}$\mathbf{k}$ metric (\S \ref{sec:setup}), which is widely adopted to study capability boundary \citep{yue2025does}. It aligns with our objective of characterizing the \textit{potential} of reasoning for factual recall, and not only the current models' top-1 behavior, since it emphasizes the presence of successful reasoning paths in the model’s output distribution while being less sensitive to their exact ranking.

\textbf{Reasoning Unlocks Latent Knowledge.} \Cref{fig:on_vs_off} compares pass@$k$ curves between reasoning \texttt{ON} and \texttt{OFF}. We use a maximum of $N=100$ samples to explore functional reasoning paths that have reasonable likelihood but not necessarily generated in a single attempt. Reasoning consistently increases the pass@$k$ values across all considered models and datasets. While there is a tangible improvement in standard accuracy (pass@$1$), the benefit is frequently more pronounced at higher $k$ values; in some cases pass@$k$ nearly doubles with reasoning (e.g., Qwen3-32B on \texttt{SimpleQA-Verified}). This persistent (and often widening) gap suggests that reasoning expands the model's parametric recall boundary. This closely resonates with findings from \cite{gekhman2025insideout}, who showed that LLMs can encode a fact yet fail to generate it even after large-scale repeated sampling, and suggests that reasoning acts as a mechanism that helps the model to better expose its internal knowledge.

\textbf{Reasoning Effectiveness Measure: A Weighted Average Improvement.} While pass@$k$ curves provide granular detail, we would also like to report a single summary metric for an overall estimate of the effect of reasoning in each setting.
To this end, we define a unified reasoning effectiveness metric $\Omega$ that accounts for the entire range of $k$ values. We identify two key requirements for $\Omega$: it must (1) measure the effect \textit{relative} to reasoning \texttt{OFF}, and (2) assign higher importance to larger $k$ values. 
The latter aims to capture our focus on capability boundary, rewarding models where the \texttt{ON} performance \textit{diverges} from \texttt{OFF} as $k$ increases.\looseness=-1

For each $k$ from $1$ to $N$ ($N=100$ in our study), we calculate the \textit{relative} percentage difference in pass@$k$ between reasoning \texttt{ON} and \texttt{OFF} modes. We then define $\Omega(N)$ as a weighted average of these differences, assigning higher importance to larger $k$ values via a linear weight $k$:\looseness=-1

\begin{equation}
    \Omega(N) = \sum_{k=1}^{N} \Bigg[ \underbrace{\textcolor{blue}{k}}_{\substack{\textcolor{blue}{\text{Linear Boundary}} \\ \textcolor{blue}{\text{Weight}}}} \cdot \underbrace{ \frac{\text{pass}@k_{\text{ON}} - \text{pass}@k_{\text{OFF}}}{\text{pass}@k_{\text{OFF}}} }_{\substack{\text{ Improvement in pass@k} \\ \text{relative to OFF baseline}}} \Bigg] \cdot \underbrace{\textcolor{blue}{\frac{1}{\sum_{k'=1}^{N} k'}}}_{\substack{\textcolor{blue}{\text{Normalization by}} \\ \textcolor{blue}{\text{sum of weights}}}}
    \label{eq:omega}
\end{equation}

\begin{wrapfigure}[15]{r}{0.45\columnwidth} % [11] forces it to only wrap 11 lines
    \vspace{-1.5em} % Pulls the entire figure upwards
    \centering
    \includegraphics[width=\linewidth]{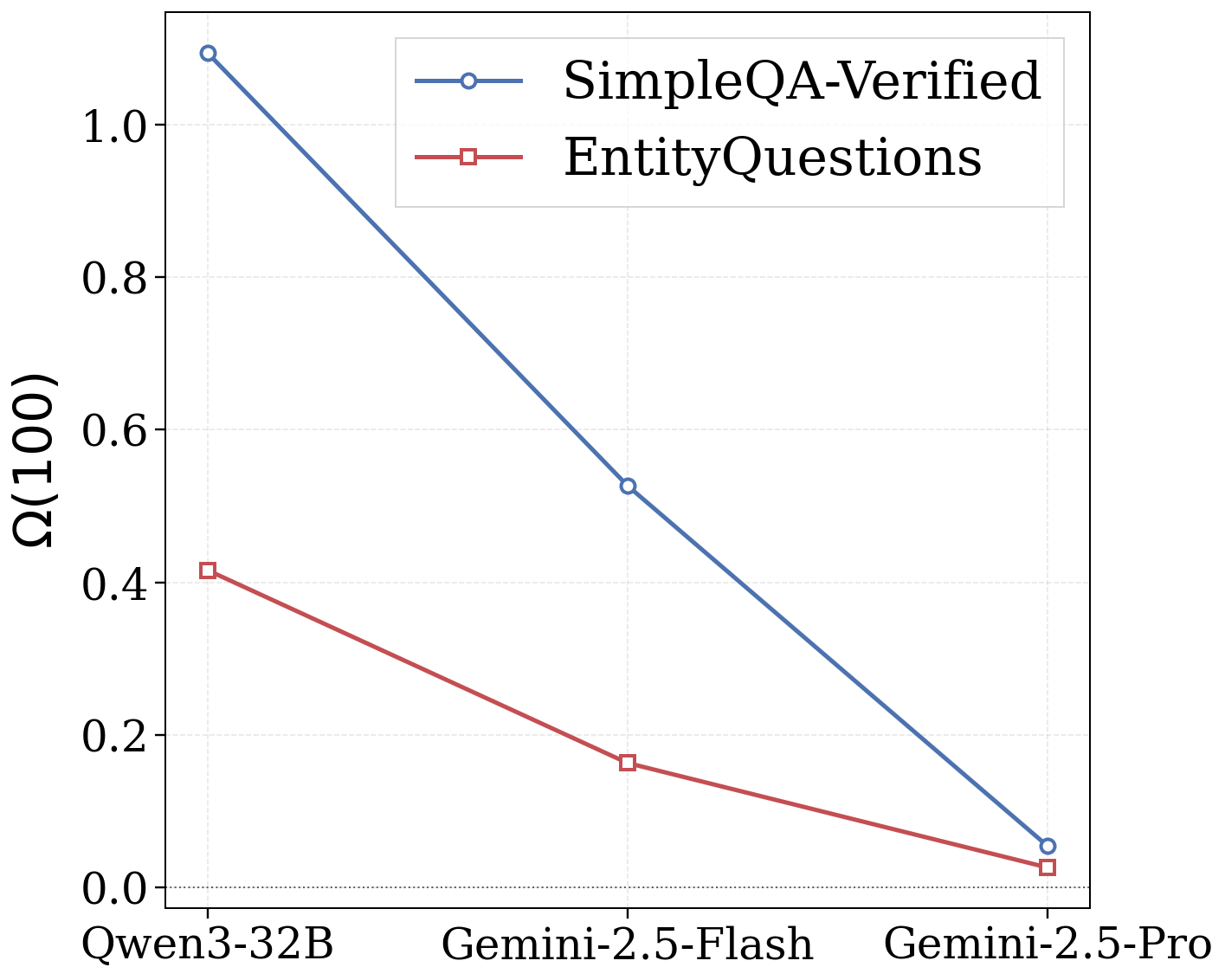}
    \caption{
    $\Omega$ in all models and datasets (\S \ref{sec:setup}). Models organized from the most (left) to the least effective (right) in terms of pass@$1$.}   
    % Reasoning effectiveness ($\Omega$) as a function of model capability on \texttt{SimpleQA-Verified} and EntityQuestion datasets.}
    \label{fig:resoning_effectiveness}
    \vspace{-1em} % Reduces the invisible padding below the caption
\end{wrapfigure}

\textbf{Less Capable Models Benefit More from Reasoning.} \Cref{fig:resoning_effectiveness} illustrates how $\Omega$ decreases as model capability increases, suggesting that more capable models are more effective at utilizing their parametric knowledge without reasoning. The high $\Omega$ observed in less capable models (e.g., Qwen3-32B) implies they possess more ``hidden knowledge'' \citep{gekhman2025insideout}, and that reasoning compensates for ineffective parametric knowledge recall. We also observe a consistently higher $\Omega$ on SimpleQA compared to \texttt{EntityQuestions}. We attribute this to the lower baseline (\texttt{OFF}) performance on SimpleQA, which implies that the specific facts targeted by this dataset are harder to fetch from the model's parameters. Consequently, there is more headroom for reasoning to compensate for ineffective recall.

\section{Question Complexity is a Poor Predictor of Reasoning Effectiveness}
\label{sec:question_complexity}

\begin{wrapfigure}[18]{r}{0.45\columnwidth}
    \vspace{-1.0em} % Pulls the entire figure upwards
    \centering
    \includegraphics[width=\linewidth]{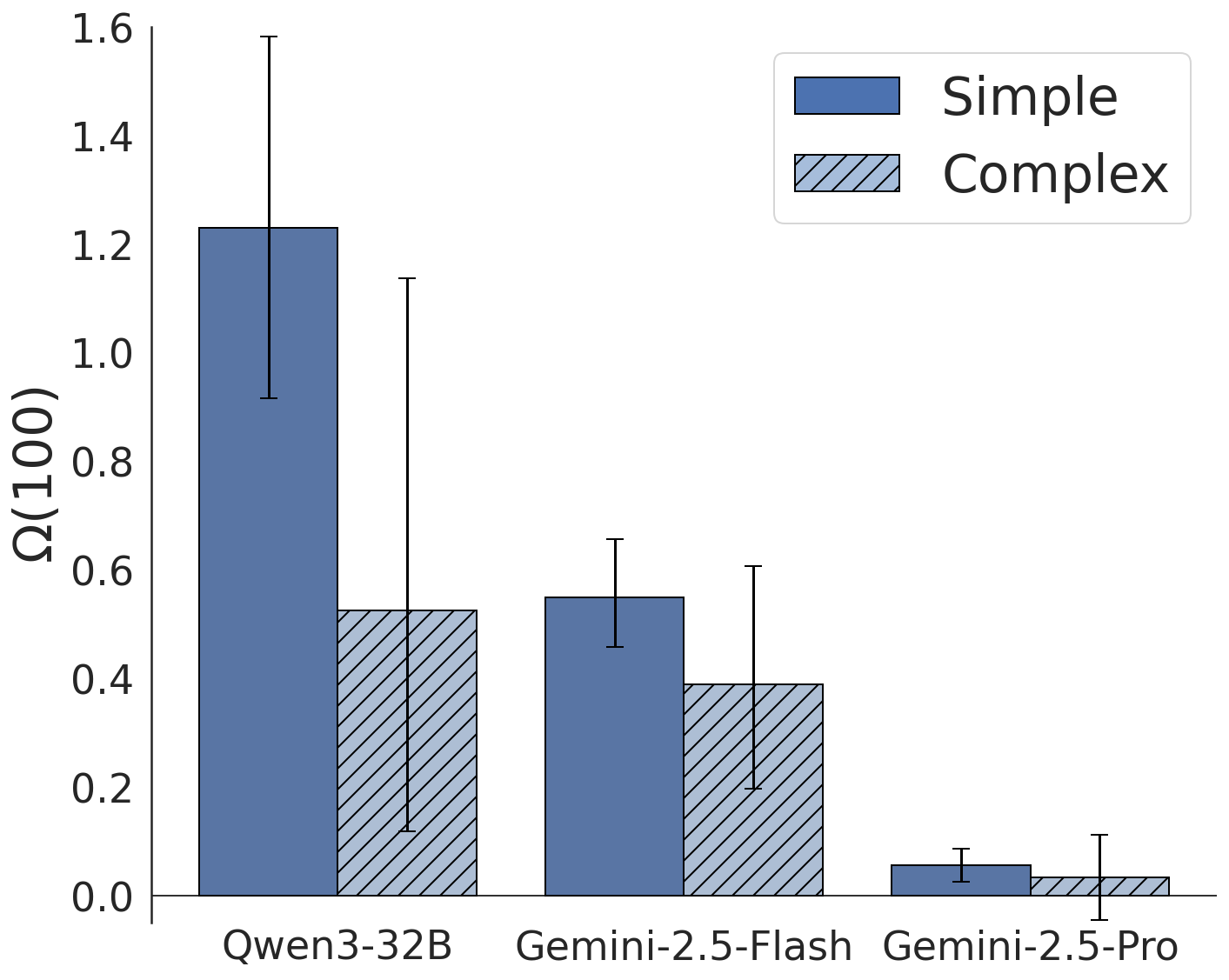}
    \caption{Reasoning Effectiveness on different question types in \texttt{SimpleQA-Verified}, with 95\% confidence intervals.}
    \label{fig:question_difficulty}
\end{wrapfigure}

A natural hypothesis for the utility of reasoning is that it aids in the decomposition of complex, multi-hop questions \citep{press-etal-2023-measuring}.
We emphasize, however, that this mechanism is unlikely to be the primary driver in our setup, as our datasets consist predominantly of simple, direct (single-hop) factual questions. Specifically, \texttt{EntityQuestions} is based on predefined single-hop templates (\Cref{tab:relations}), and \texttt{SimpleQA-Verified} contains metadata indicating that 90\% of questions (903 out of 1,000) are single-hop.
Nevertheless, the presence of this metadata in \texttt{SimpleQA-Verified}, combined with our experimental setup, provides an interesting lens for analyzing the impact of reasoning on simple versus complex questions.\looseness=-1

The \texttt{SimpleQA-Verified} questions' metadata indicates whether they \textit{require reasoning} and whether they are \textit{multi-step}. We define \texttt{Complex} questions as those for which at least one of the these labels is True, and term the remaining questions \texttt{Simple}. \Cref{fig:question_difficulty} compares the reasoning effectiveness $\Omega$ (Equation \ref{eq:omega}) between these two subsets. Perhaps contrary to expectation, we find no evidence that reasoning yields higher marginal gains on the \texttt{Complex} subset, as the 95\% confidence intervals of the two subsets are overlapping. For Gemini-2.5-Pro, the benefit of reasoning is not even guaranteed for the \texttt{Complex} subset, as the 95\% confidence interval crosses zero. We emphasize that this analysis does not fully isolate complexity: the number of complex examples is relatively small (leading to large confidence intervals), and proper isolation would require constructing complex and simple variants of the same underlying question. Nevertheless, even under these limitations, the absence of a stronger effect on the \texttt{Complex} subset, explicitly labeled as requiring reasoning, is surprising. It reinforces the evidence that, in our study, the gains are not primarily driven by task decomposition, but from facilitating parametric recall. This naturally raises the question of what mechanisms enable reasoning to improve recall.

\section{\textit{How} Reasoning Improves Parametric Recall?}
\label{sec:analysis}

After establishing that gains from reasoning mostly stem from aiding parametric recall rather than task decomposition (\S\ref{sec:question_complexity}), we analyze the mechanisms behind this effect. We adopt a hypothesis-driven approach in which we formulate candidate explanations and then design controlled experiments to test them. We first evaluate a content-independent explanation: the \textbf{computational buffer} hypothesis (\S\ref{sec:test_time_compute}), under which the extra reasoning tokens allow to perform additional \textit{latent} computation before generating the answer. We next analyze the \textit{content} of the reasoning traces. Since simple questions do not require step-by-step derivations, the traces rarely contain such solution paths. Instead, they often surface facts that are topically related to the question. This motivates the \textbf{factual priming} hypothesis (\S \ref{sec:factual_priming}): generating relevant factual context before answering facilitates retrieval of the correct answer. Since factual priming relies on surfacing factual content, a natural follow-up question concerns its \emph{correctness}. We thus examine the role of \textbf{hallucinations} in the trace and how they affect final answer (\S \ref{sec:hallucinations}). Since our controlled experiments are compute heavy, we focus on the model with the best latency-quality tradeoff which is \texttt{Gemini-2.5-flash} and run the experiments on both \texttt{SimpleQA-Verified} and \texttt{EntityQuestions}. We conclude the section by showing that our insights can be harnessed to improve model accuracy by prioritizing reasoning trajectories that exhibit the desired traits (\S \ref{sec:improving_accuracy}).\looseness=-1

\subsection{Reasoning Tokens as a Computational Buffer}
\label{sec:test_time_compute}

\begin{figure*}[t] 
    \centering
    \includegraphics[width=0.75\textwidth, keepaspectratio]{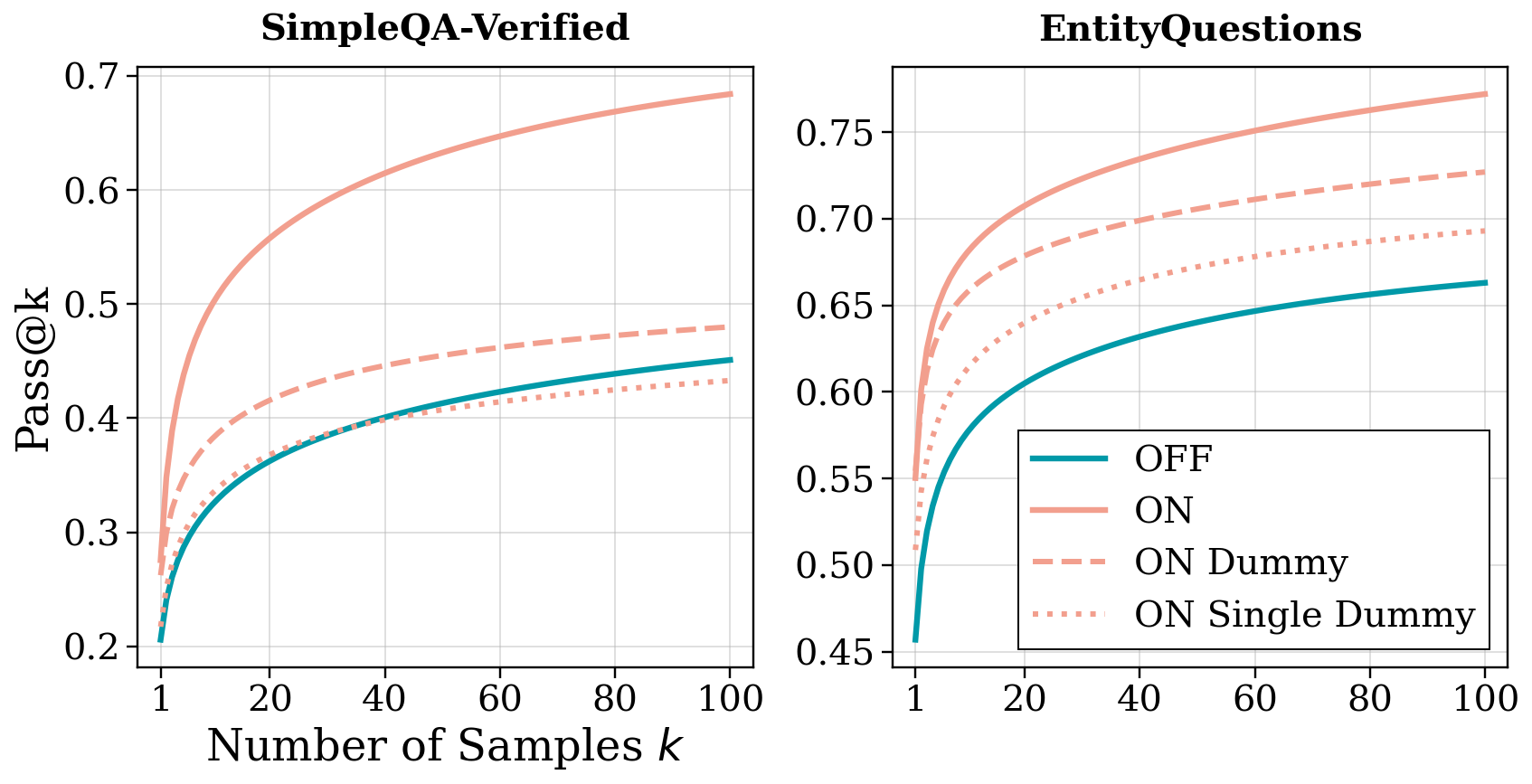}
    \caption{Computation buffer effect on Gemini-2.5-Flash (\S \ref{sec:test_time_compute}). 
    \texttt{ON\,Single\,Dummy} overrides the thinking trace with a short dummy sequence. \texttt{ON\,Dummy} does the same, but repeats the short dummy sequence to match the length of the original trace.}
    \label{fig:dummy_thought}
\end{figure*}

In this section we explore the computational buffer hypothesis: generating extra tokens during reasoning allows models to perform additional latent operations and to bypass the depth limits of a single forward pass on the input \citep{goyal2024think}. While prior work discussed this hypothesis, it was not directly tested in modern R-LLMs, nor examined in the context of parametric knowledge recall. We hypothesize that R-LLMs implicitly exploit this mechanism to retrieve hard-to-reach parametric knowledge. To isolate the effect of equivalent computation without conditioning on the content of the original trace, we introduce the \texttt{ON\,Dummy} variant, in which we replace the model’s original reasoning trace with the semantically meaningless dummy sequence \textit{``Let me think.’’} repeated to match the original trace's length (see example in \Cref{fig:on_dummy_case_study}), and then regenerate the final answer conditioned on it. As shown in \Cref{fig:dummy_thought}, conditioning on this meaningless trace substantially improves pass@$k$ over \texttt{OFF}. The gains also translate to pass@$1$: accuracy rises from $0.206$ to $0.262$ on \texttt{SimpleQA-Verified} and from $0.457$ to $0.554$ on \texttt{EntityQuestions}.

We next rule out a potential confounder, which we term \textbf{ON/OFF bias}: the model may perform better in \texttt{ON} not because of extended computation, but due to a training-induced preference for that mode — for instance, if its training data contains more \texttt{ON} examples than \texttt{OFF}. To control for this, we introduce \texttt{ON\,Single\,Dummy}, which is identical to \texttt{ON\,Dummy} except that the dummy string appears only once rather than being repeated to match the original trace length. Both variants have no semantic content in their reasoning trace and operate in \texttt{ON} mode; the only difference is computational length. The consistent performance gap between them (\Cref{fig:dummy_thought}) therefore isolates the effect of additional computation, providing strong evidence that the model leverages it to refine its predictions. In \S \ref{appendix:case_study_compute} we present a case study where pure compute aids the model in recalling the correct answer.

 \begin{figure*}[t] 
    \centering
    \includegraphics[width=0.75\textwidth, keepaspectratio]{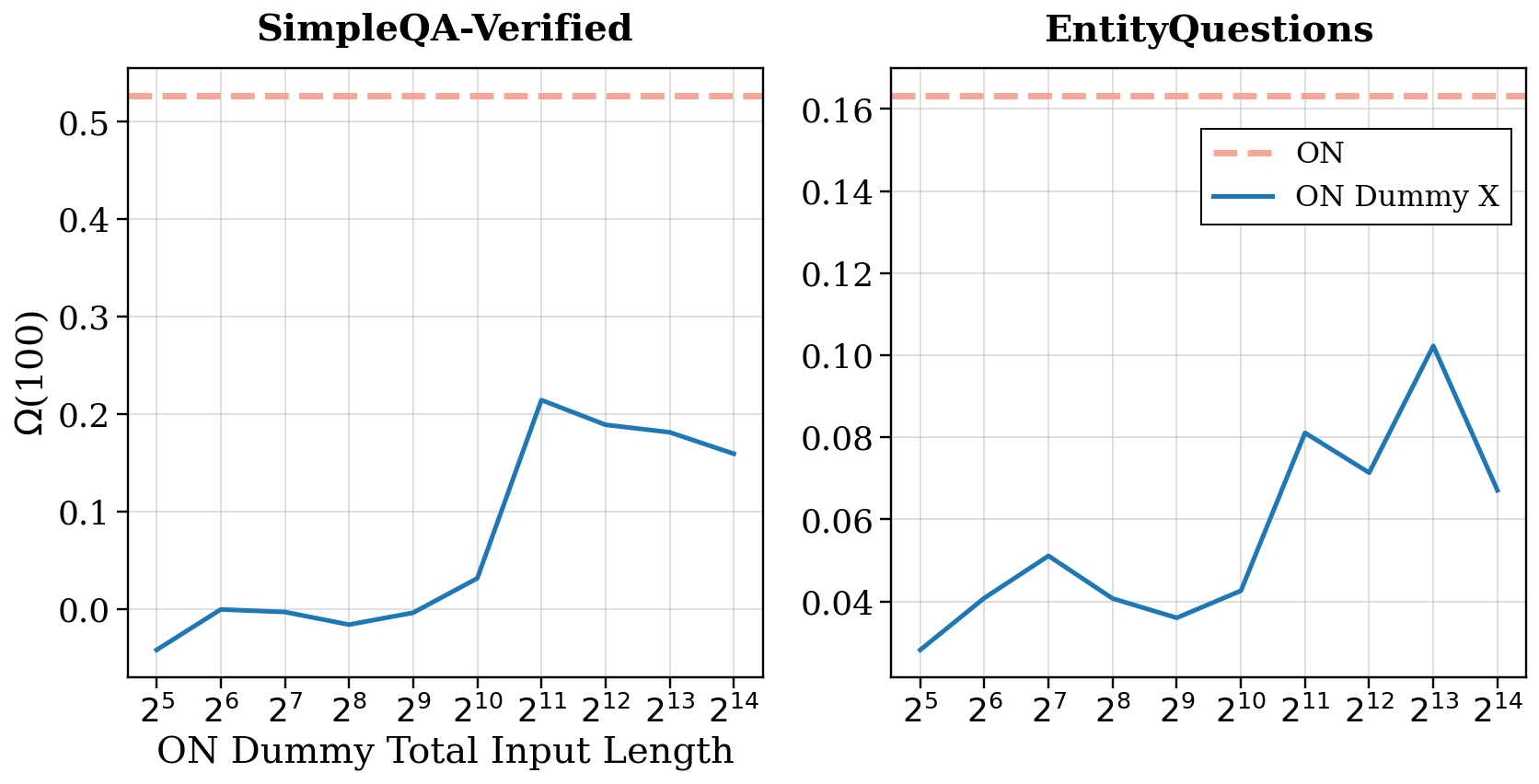}
    \caption{
    % The effect of increasing compute. 
    Reasoning effectiveness (Equation \ref{eq:omega}) as a function of the input length in tokens when conditioning on dummy reasoning trace (see \S \ref{sec:test_time_compute}).
    \texttt{ON\,Dummy\,X} overrides the reasoning trace with a short dummy sequence which is repeated such that the input length will be X.}
    \label{fig:dummy_scaling_omega}
    \vspace{-5pt}
\end{figure*}

Finally, we test if more computation is always better. While prior work showed that longer reasoning traces are not necessarily beneficial \citep{yang2025towards,hassid2025don,feng2025characterizes}, they focus on classical reasoning tasks such as math rather than factual recall, and do not isolate pure computation, leaving it unclear whether shorter traces help because of their semantic content or their reduced computational budget. In \Cref{fig:dummy_scaling_omega}, we vary the length of the dummy trace and observe a non-monotonic scaling pattern. We used our aggregated reasoning effectiveness measure (Equation \ref{eq:omega}) for clear visualization
of the trends. Increasing the dummy length initially shifts the pass@$k$ curves upward, but saturates or even becomes counterproductive beyond a certain point. For example, on \texttt{SimpleQA-Verified}, pass@k improves as the dummy length grows up to 2048 tokens ($2^{11}$) (despite minor fluctuations), yet extending it to 4096 ($2^{12}$), 8192 ($2^{13}$), and 16384 ($2^{14}$) tokens leads to a consistent decline. We provide the full pass@$k$ curves for completeness in  \Cref{fig:dummy_scaling} in the Appendix.\looseness=-1 

Collectively, these results show that R-LLMs can latently use additional computation during reasoning-token generation, independent of trace semantics. However, this effect is bounded, and scaling dummy computation never fully recovers the performance of reasoning \texttt{ON}, suggesting that the compute-buffer mechanism alone cannot account for all observed gains, motivating a closer examination of the semantic content of reasoning traces.

\subsection{Factual Priming}
\label{sec:factual_priming}

An initial qualitative inspection of the contents of the reasoning traces, confirmed that, as expected, they rarely contain multi-step logical derivations. Instead, they tend to list candidate answers, recall relevant facts, or describe an intended search process.\footnote{Although search is disabled at evaluation, reasoning traces often contained search plans.} Considering this, the most relevant content we identify is mentioning related facts. 

In human cognition, processing a concept spreads ``activation'' through a semantic network, priming related neighbors by lowering the threshold for their retrieval \citep{collins1975spreading}. We hypothesize that R-LLMs exhibit a similar, cognitively plausible mechanism where the model engages in generative self-retrieval, effectively constructing a contextual bridge to the answer through recalling related facts, and term this mechanism \textbf{factual priming}.

To test this hypothesis, we extract a list of facts mentioned in the reasoning trace using a prompted LLM. This extraction step presents several challenges. First, traces can restate information from the question. Since we aim to capture facts introduced during reasoning, we remove such restatements from the extracted list with an additional LLM-based step. Second, the model may commit to an answer during reasoning, explicitly stating its intention to output it. We therefore apply an additional filtering step that removes answer-disclosing statements with respect to both (i) the gold answer and (ii) the model’s predicted answer, which may differ. Importantly, removing every occurrence of the answer would be overly aggressive, as the model may mention it as part of recall rather than as a committed resolution. We therefore remove only statements that explicitly link the target answer to the question. Full technical details are provided in \S\ref{appendix:facts_extraction}.

To test the contribution of recalled facts, we create two variants both of which condition the prediction on the extracted fact list. \texttt{ON\,Facts}, overrides the reasoning trace with the extracted fact list and re-generates the answer. In \texttt{OFF\,Facts}, the model runs with reasoning \texttt{OFF} while the facts list is provided as additional context (see example in \Cref{fig:off_summary_case_study}). \texttt{OFF\,Facts} aims to control for confounders from running with reasoning enabled. For example, \texttt{ON\,Facts} may push the model beyond its typical distribution, as it is not optimized to operate on traces that look like a fact list. In contrast, the model is optimized for instruction following and should be able to use additional input context. These variants cannot be simply compared to reasoning \texttt{OFF}, mainly since they both use more test-time compute (\S\ref{sec:test_time_compute}). To control for this, we introduce \texttt{OFF\,Dummy\,Facts} and \texttt{ON\,Dummy\,Facts}, which replace the facts with a semantically meaningless dummy strings of similar length (more details in \S \ref{appendix:off_facts}).

\begin{figure*}[t] 
    \centering
    \includegraphics[width=0.75\textwidth, keepaspectratio]{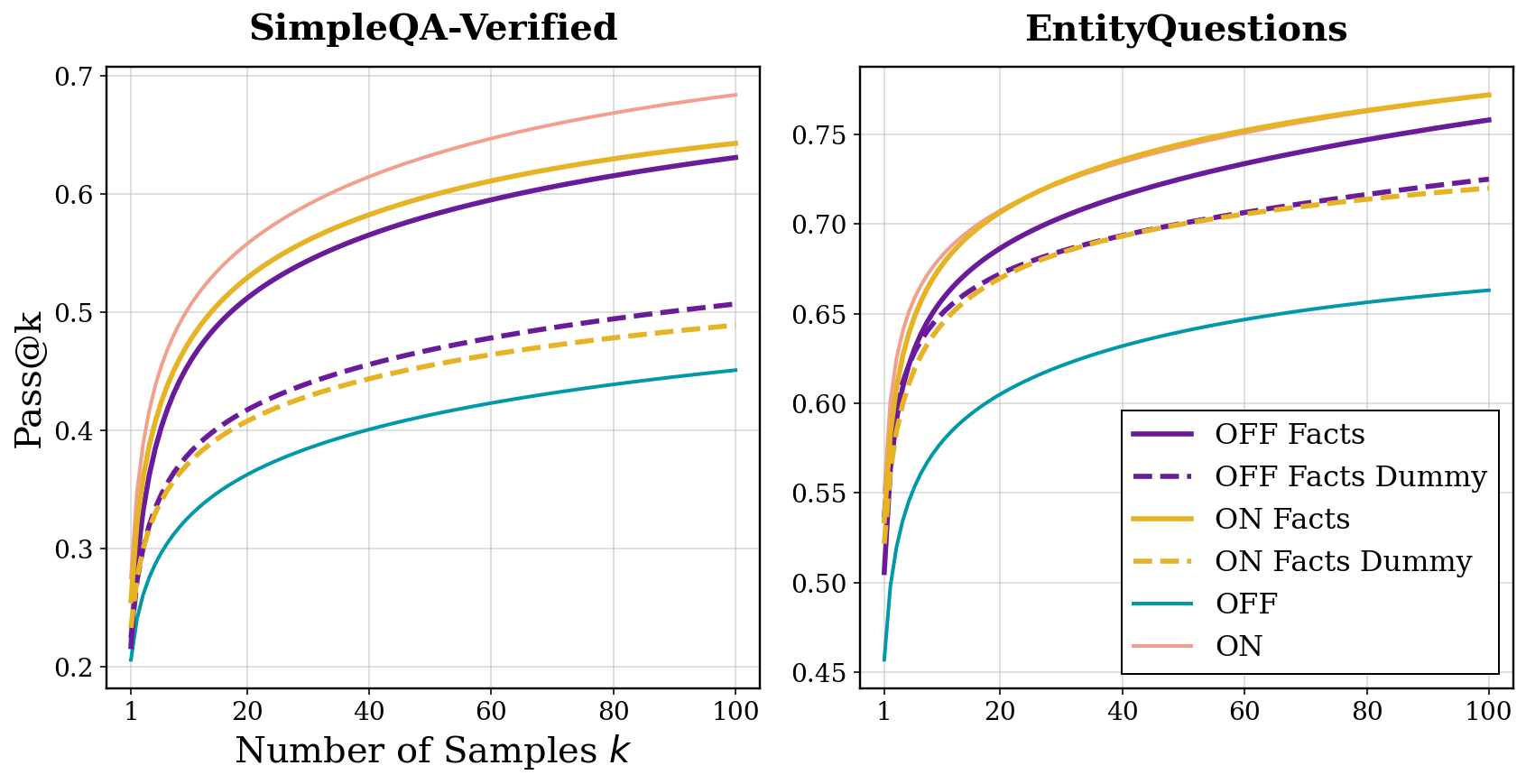}
    \caption{
    Factual priming effect on Gemini-2.5-Flash (\S\ref{sec:factual_priming}), conditioning on facts recalled during reasoning, with reasoning either \texttt{OFF} (\texttt{OFF\,Facts}) or \texttt{ON} (\texttt{ON\,Facts}).}
    \label{fig:off_summary_plot}
\end{figure*}

\Cref{fig:off_summary_plot} presents the results. Both baselines considerably outperform their dummy variants, providing strong evidence in favor of the factual priming hypothesis. Interestingly, facts help even when we disable reasoning (\texttt{OFF\,Facts}), which reinforces the conclusion that the facts themselves are useful for answer recall. \texttt{ON\,Facts} performs even better than \texttt{OFF\,Facts}, suggesting that the model is relatively robust to the aforementioned distribution shift and may exhibit a positive bias toward the \texttt{ON} mode (see \S \ref{sec:test_time_compute}). For \texttt{EntityQuestions}, \texttt{ON\,Facts} even matches the performance of \texttt{ON} while using dramatically less compute. In \S \ref{appendix:case_study_priming} we present a case study where the facts aid the model in recalling the correct answer.

\subsection{Reasoning Hallucinations Encourage Final Answer Hallucinations}
\label{sec:hallucinations}

While recalling related facts during reasoning can facilitate correct answers (\S \ref{sec:factual_priming}), such recall is not guaranteed to be accurate. LLMs are prone to hallucinations, and this vulnerability extends to facts generated during reasoning. To test if such hallucinations increase the likelihood of hallucinated final answers, we detect hallucinated facts within reasoning traces using a prompted LLM with access to web search. For each question and each reasoning ON sample ($100$ samples per-question), we use the previously extracted facts from the reasoning trace (\S \ref{sec:factual_priming}) and verify each \textit{independently} using Gemini-2.5-Flash with search enabled, prompting it to assess correctness and allowing abstention if correctness cannot be reliably determined (e.g., vague or underspecified statements). We validate this procedure through human evaluation on a small set: among the facts ultimately classified as correct or incorrect (excluding abstentions), the estimated verification accuracy is $\sim\!100\%$. Additional implementation details are provided in \S \ref{appendix:detecting_hallucinations}.\looseness=-1

We discard traces without facts, and traces whose facts could not be verified (where the search-based verifier abstained), ensuring that the remaining traces contain only facts with reliable correctness labels. The remaining traces are labeled as \emph{clean} if all intermediate facts are correct, and as \emph{hallucinated} if they contain at least one incorrect fact. When pooling traces across all questions, we observe a large and consistent gap: hallucinated traces are substantially less likely to produce correct final answers. On \texttt{SimpleQA-Verified}, 41.4\% of clean traces yield correct answers, compared to 26.4\% of hallucinated traces. A similar gap appears on \texttt{EntityQuestions}, where the proportion of correct final answers drops from 71.1\% for clean traces to 32.2\% for hallucinated traces.

\begin{figure*}[t] 
    \centering
    \includegraphics[width=0.7\textwidth, keepaspectratio]{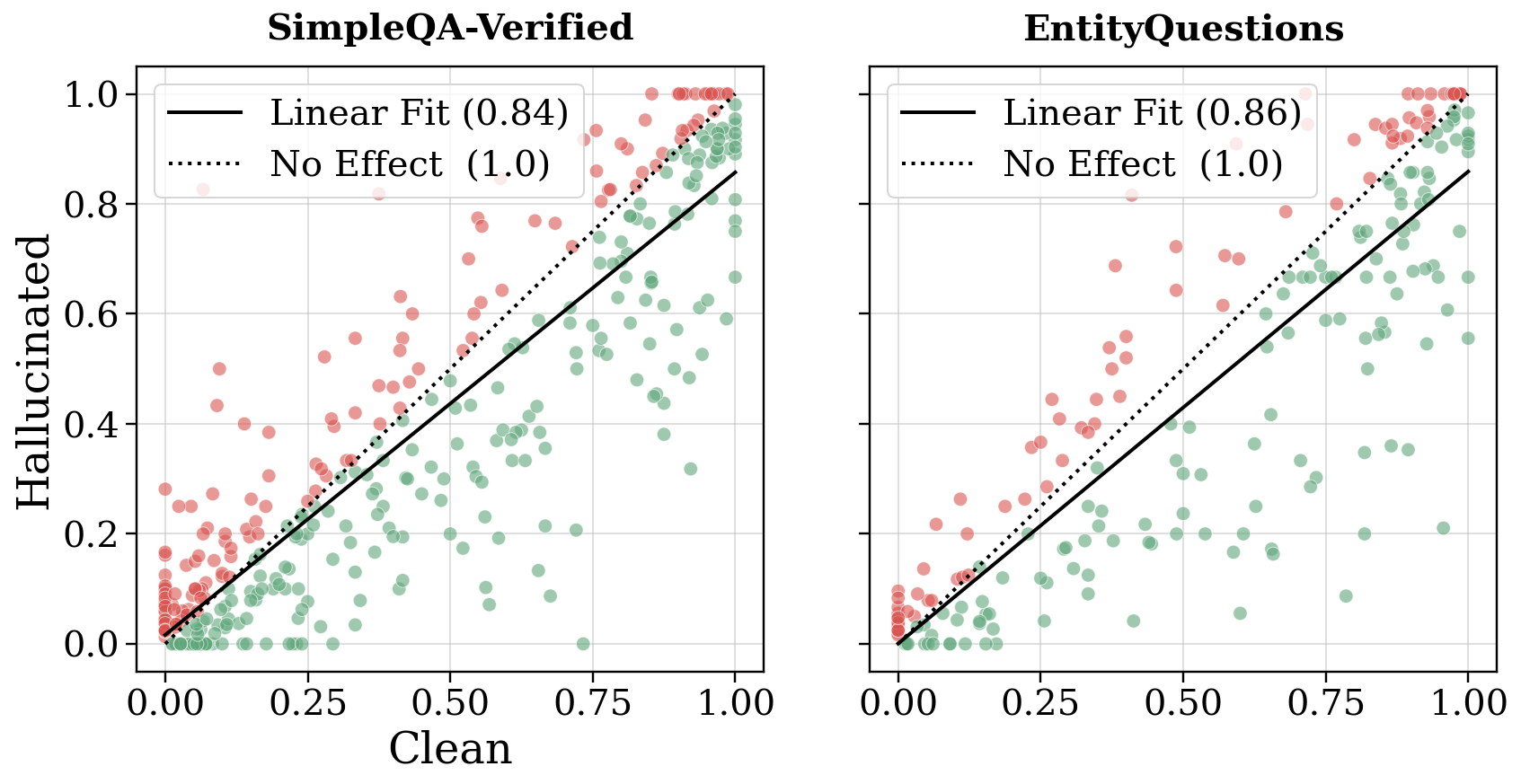}
    \caption{
    Within-question comparison of correct final-answer rates in clean (x-axis) vs. hallucinated (y-axis) reasoning traces. Each question is one point; red examples lie above the no-effect diagonal, green below.}
    \label{fig:hallucinations_within_question}
\end{figure*}

Although the aggregate analysis reveals a strong signal, it does not control for question difficulty: harder questions may simultaneously induce lower final-answer correctness and more hallucinated intermediate facts. To disentangle these effects, we compare the proportion of correct final answers between the clean and hallucinated subsets \textit{within each question}. To obtain a reliable signal, we only use questions for which each subset contains at least 10 traces. We exclude uninformative questions where final answer correctness is identical across subsets—either 0\% in both (consistently hard) or 100\% in both (consistently easy).
\Cref{fig:hallucinations_within_question} shows, for each question, the proportion of correct final answers in the clean subset versus the hallucinated subset. The fitted regression line has a slope below 1 (0.84 for \texttt{SimpleQA-Verified} and 0.86 for \texttt{EntityQuestions}), indicating a systematic within-question gap: even after controlling for question difficulty, traces that contain hallucinated intermediate facts are less likely to produce correct final answers.

\subsection{From Analysis to Practice: Improving Accuracy by Sampling High-Potential Traces}
\label{sec:improving_accuracy}

We showed that reasoning improves parametric knowledge recall through two mechanisms: additional computation (\S \ref{sec:test_time_compute}) and factual priming (\S \ref{sec:factual_priming}), and that hallucinated intermediate facts decrease the probability of a correct final-answer (\S \ref{sec:hallucinations}). We now ask whether these insights can be translated into practical accuracy gains. The computational buffer effect does not provide a reliable control signal, as longer traces are not consistently better and the optimal length is unknown. In contrast, factual signals suggest a concrete opportunity: we can detect whether a trace recalls explicit facts and whether those facts are correct, and prioritize traces that surface verified facts.

\begin{table}[h]
    \centering
    % The first argument {0.5\textwidth} matches the wraptable width
    % The second argument {!} maintains the aspect ratio
    \resizebox{0.6\textwidth}{!}{
        \begin{tabular}{@{}lll@{}}
        \textbf{Strategy} & \textbf{SimpleQA-Verified} & \textbf{EntityQuestions} \\ \toprule
        Regular           & 27.9                       & 56.9                     \\
        Only Facts        & 30.2 (+8.2\%)              & 58.4 (+2.6\%)            \\
        Only Correct Facts & \textbf{31.3 (+12.2\%)}    & \textbf{59.8 (+5.1\%)}   \\ \bottomrule
        \end{tabular}
    }
    \caption{Expected accuracy under test-time selection criteria based on factual recall and factual correctness.
    % Relative improvements from selecting traces that (i) recall facts and (ii) recall only verified facts.
    }
    \label{tab:improving_accuracy}
\end{table}

Since training the model to prioritize such traces during generation is non-trivial, we adopt a test-time selection strategy to probe the gains implied by our findings. We consider two selection criteria: (i) retaining traces that explicitly recall facts, and (ii) retaining traces that both recall facts and contain no hallucinated intermediate facts. Treating the LLM as a generation oracle, we simulate the expected accuracy under each criterion using multiple samples per question (technical details in \S \ref{appendix:hallucinations_simulation}). \Cref{tab:improving_accuracy} presents the results. Selecting traces that recall facts yields relative improvements of 8.2\% and 2.6\% on \texttt{SimpleQA-Verified} and \texttt{EntityQuestions}, respectively. Further restricting to traces whose recalled facts are correct increases these gains to 12.2\% and 5.1\%. These results suggest that explicitly favoring on-policy reasoning traces that contain factual statements and avoid hallucinations may be a promising direction for improving factual accuracy. Such prioritization could be implemented through training with process rewards \citep{lightman2024lets} that encourage factually supported intermediate steps.

\section{Case Studies}
\label{appendix:case_studies}

 \begin{figure*}[t!] 
    \centering
    \includegraphics[width=0.8\textwidth, keepaspectratio]{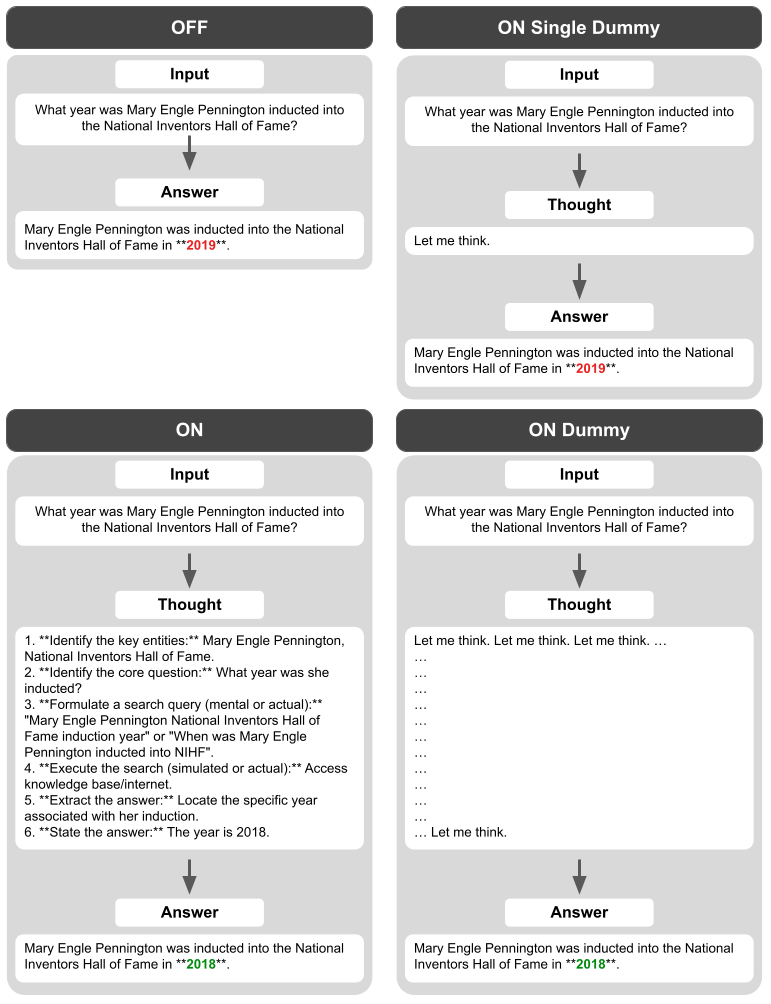}
    \caption{
    Case study for the effectiveness of the computational buffer effect.}
    \label{fig:on_dummy_case_study}
    % \vspace{-5pt}
\end{figure*}

\subsection{The Computational Buffer Effect}
\label{appendix:case_study_compute}

\Cref{fig:on_dummy_case_study} presents a case study in which Gemini-2.5-Flash benefits from a \textit{dummy} reasoning trace. The question is \textit{``What year was Mary Engle Pennington inducted into the National Inventors Hall of Fame?''} and the model fails to answer it without reasoning (\texttt{OFF}) and predicts the (wrong) year \textit{``2019''}. When reasoning is enabled, the model successfully predicts the correct year (\textit{``2018''}). Interestingly, the reasoning trace (Thought) does not seem to contain any useful information apart form restating the information from the question and planning to execute search, which suggest that the real benefit from reasoning in this case may come simply from additional compute. Supporting this interpretation, \texttt{ON\,Dummy}, which overrides the reasoning trace with a dummy string with similar length, also leads to correct answer. This is not the case with \texttt{ON\,Single\,Dummy}, which overrides the trace with a very short dummy string and leads to a wrong prediction, which further reinforces the conclusion that the model needed to perform additional computational steps to recall the correct answer.

\subsection{Factual Priming}
\label{appendix:case_study_priming}

 \begin{figure*}[t!] 
    \centering
    \includegraphics[width=0.8\textwidth, keepaspectratio]{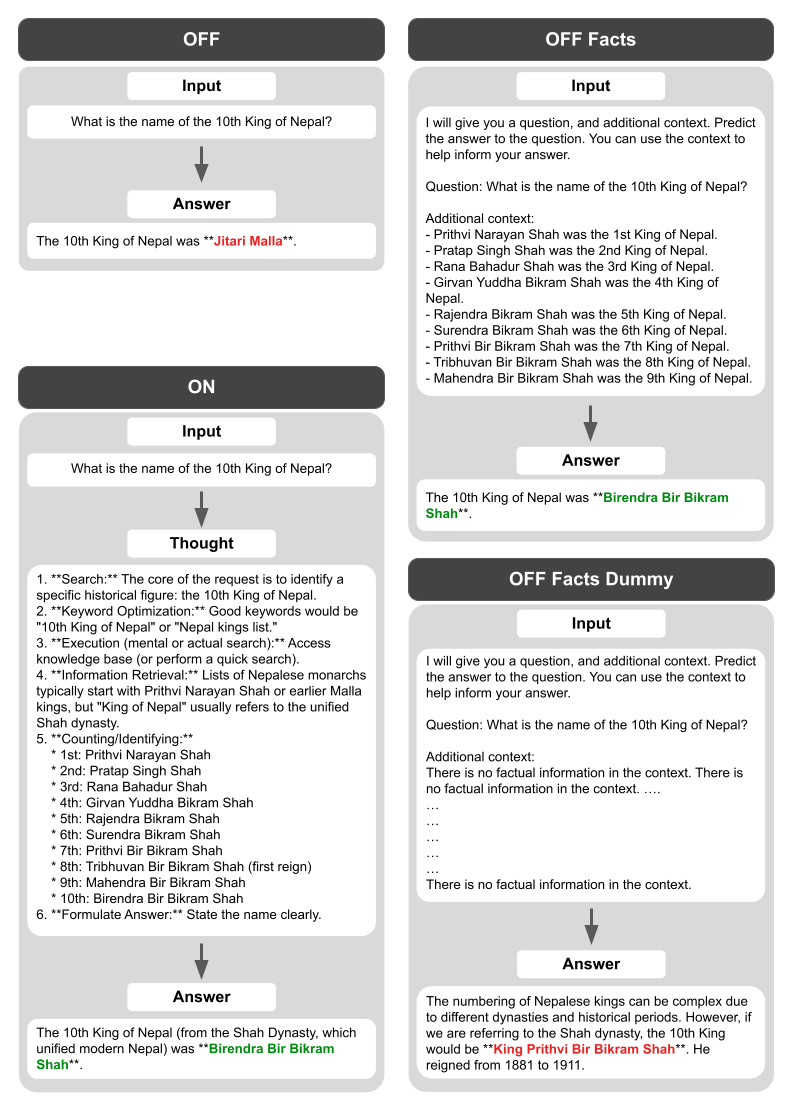}
    \caption{
    Case study for the effectiveness of factual priming.}
    \label{fig:off_summary_case_study}
    % \vspace{-5pt}
\end{figure*}

\Cref{fig:off_summary_case_study} presents a case study in which Gemini-2.5-Flash benefits from factual priming. The question is \textit{``What is the name of the 10th King of Nepal?''} and the model fails to answer it without reasoning (\texttt{OFF}) and predicts the (wrong) answer \textit{``Jitari Malla''}. When reasoning is enabled, the model successfully predicts the correct year (\textit{``Birendra Bir Bikram Shah Dev''}).

The reasoning trace (Thought) contains many related facts, listing all the 10 kings. When we extract these facts, disable reasoning and provide them as additional context to the model (\texttt{OFF\,Facts}), the model also successfully recalls the correct answer. In addition, \texttt{OFF\,Dummy\,Facts}, which matches the compute of \texttt{OFF\,Facts} but uses a dummy string instead of the facts, fails to predict the correct answer, predicting \textit{''King Prithvi Bir Bikram Shah''} which is the name of the 7th king and not the 10th. This suggests that the success in \texttt{OFF\,Facts} is not due to the increased compute compared to \texttt{OFF}. 
Importantly, since during facts extraction we extract only facts that do not reveal the answer, the facts that \textit{``Birendra Bir Bikram Shah''} was the 10th king that is mentioned in the reasoning traces is removed from the list. This case provide a classic example of factual priming. We showed that the facts themselves are useful and we showed that recalling the first 9 kings helps the model to recall the 10th.

\section{Related Work}

\textbf{Reasoning in Closed-Book QA.} Reasoning for parametric knowledge recall has been extensively studied for complex, multi-step questions, where intermediate steps are naturally useful for making gradual progress towards the solution \citep[\textit{inter-alia}]{press-etal-2023-measuring,yao2023iclr,trivedi-etal-2023-interleaving}. However, it is less clear if the gains in such settings arise from question decomposition or also reflect an improvement in knowledge recall. Two concurrent studies explored simple questions and showed that reasoning can increase \textit{accuracy} on them \citep{ma2026improving, calderon2026empty}. In this work we (1) do not limit our study to accuracy, but study the model's capability boundary (see below) and (2) provide detailed analysis for which mechanisms make reasoning helpful for parametric recall.

\textbf{Studying LLMs' Capability Boundary.} We study how reasoning affects capability boundary (whether it unlocks otherwise unreachable correct answers) using the pass@$k$ metric, which has become standard for this purpose \citep{yue2025does,guo2025deepseek,shao2024deepseekmath}. Prior capability boundary research focuses on math and code, typically comparing a base (non-reasoning) model prompted with Chain-of-Thought, to its variant that has been RL-tuned for reasoning \citep{yue2025does,shao2024deepseekmath,he-etal-2025-rewarding}. They usually report gains at small $k$ but not at large $k$, indicating probability sharpening of already-accessible answers. In contrast, we study the effect of reasoning itself and how enabling it at inference time affects the capability boundary. We show substantial gains at high $k$, indicating an expansion of the capability boundary for parametric knowledge recall.

\textbf{Reasoning as a Computational Buffer.} The computational buffer hypothesis (\S\ref{sec:test_time_compute}) was first discussed in early work on traditional Chain-of-Thought reasoning. A standard way to test it, adopted both in prior work and in our study, is replacing the reasoning trace with uninformative filler text and compare performance to a non-reasoning baseline. Early studies found no effect for such substitutions \citep{wei2022nips,lanham2023measuring}, while subsequent work showed that LLMs can be \textit{explicitly trained} to use compute from filler tokens \citep{pfau2024let,goyal2024think}. Since modern R-LLMs are \textit{trained} to reason, a natural question is whether they implicitly learn to exploit reasoning tokens as computational buffers. Prior work hints at this possibility, reporting that R-LLMs may omit the steps required to derive the answer \citep{guo2025deepseek,stechly2025beyond}, or include steps that do not correspond to the model's actual reasoning mechanisms \citep{chen2025reasoning, arcuschin2025chain}. To our knowledge, no prior study has directly tested the computational buffer hypothesis in modern R-LLMs, nor examined it in the context of parametric knowledge recall.\looseness=-1

\textbf{What Makes Reasoning Effective.} A growing line of work studies which \textit{content} makes reasoning effective, often operationalizing cognitive behaviors such as review, self-verification, backtracking, and revision, and correlating them with downstream correctness \citep{feng2025characterizes,gandhi2025cognitive,yang-etal-2025-well,muennighoff-etal-2025-s1}. Another line of work examines how reasoning \textit{length} correlates with performance \citep{wu2025more,jin-etal-2024-impact}. We focus specifically on trace properties most relevant to parametric recall, namely the factual statements that appear in the trace and their correctness (\S\ref{sec:factual_priming}). Moreover, to our knowledge, prior work has not causally decoupled the compute-buffer effect (\S\ref{sec:test_time_compute}) from specific reasoning contents, while we explicitly do so in a controlled experiment.

\section{Conclusion}

We study the effect of reasoning on LLMs' parametric knowledge recall. When probing the capability boundary with pass@$k$ at large $k$, we find that reasoning consistently increases pass@$k$, suggesting that it unlocks correct answers that are effectively unreachable without it; we further show evidence that these gains stem mostly from improved parametric recall and not multi-hop question decomposition. To understand what drives this effect, we conduct hypothesis-driven controlled experiments and identify two complementary mechanisms: (i) a \emph{computational buffer} effect, where generating additional tokens enables latent computation; and (ii) \emph{factual priming}, where recalling topically related facts provides a semantic bridge that facilitates correct answer recall. At the same time, we show that hallucinated intermediate facts systematically reduce final-answer correctness. Finally, we demonstrate that our insights can be operationalized by prioritizing trajectories that surface factual statements while avoiding hallucinated intermediate facts, yielding considerable 
accuracy improvements.\looseness=-1

\bibliography{main}

\appendix

\section{Appendix}
\label{sec:appendix}

\subsection{EntityQuestions Relations}
\label{appendix:eq_rels}

\citet{gekhman2025insideout} categorized the 24 for relations in the EntityQuestions dataset \citep{Entity_Questions} according to two criteria. (1) \textit{``Hard to Guess''}, which refers to questions where the possible answer's space is large. For instance, person names are considered hard to guess, whereas professions are not, as there are relatively few professions, and the model can default to more common ones. (2) \textit{``Well Defined''}, which assesses whether the entity type and answer granularity are unambiguous. We build on this categorization and focus on the 4 relations which were found to be both hard to guess and well defined. We present these relations in \Cref{tab:relations}.

\begin{table}[h]
\centering
\resizebox{0.5\columnwidth}{!}{%
\begin{tabular}{ll}
\textbf{Relation} & \textbf{Question Template} \\ 
\toprule
P176 & Which company is [X] produced by? \\
P264 & What music label is [X] represented by? \\
P50 & Who is the author of [X]? \\
P26 & Who is [X] married to? \\
\bottomrule
\end{tabular}%
}
\caption{Overview of the relations that we use from the EntityQuestions dataset and their corresponding question templates.}
\label{tab:relations}
\end{table}

\subsection{Inference Parameters}
\label{appendix:hyperparams}
When reasoning is \texttt{ON} we set the maximum output token limit to 32,768, similarly to \citet{guo2025deepseek}. For Gemini-2.5-Flash we use identical sampling parameters for both thinking \texttt{OFF} and \texttt{ON}, setting temperature to $T=1.0$ and \textit{top-}$p$ to $=0.95$, aligning with the default values in Google AI Studio.\footnote{\url{https://aistudio.google.com/prompts/new_chat}} For Qwen3-32B, we follow the official best practices,\footnote{\url{https://huggingface.co/Qwen/Qwen3-32B\#best-practices}} setting $T=0.6$, \textit{top-}$p=0.95$, \textit{top-}$k=20$, and \textit{min-}$p=0$ for thinking \texttt{ON} and $T=0.7$, \textit{top-}$p=0.8$, \textit{top-}$k=20$, and \textit{min-}$p=0$ for thinking \texttt{OFF}.

\subsection{Simulating the Expected Accuracy With Test Time Selection}
\label{appendix:hallucinations_simulation}

As mentioned in \S \ref{sec:improving_accuracy}, we consider two selection criteria: (i) retaining traces that explicitly recall facts, and (ii) retaining traces that both recall facts and contain no hallucinated intermediate facts. For this purpose, we use our 100 generated samples per question. For each question, we create 3 subsets: (1) a subset of all samples, (2) a subset with only traces that contain facts and (3) a subset with only traces that contain facts that are all correct. We filter-out questions for which some subsets are empty, which slightly reduced the number of questions from 1,000 to 975 for \texttt{SimpleQA-Verified} and from 1,000 to 961 for \texttt{EntityQuestions}. We then check the probability of sampling the correct answer from each set per question.

\subsection{The Facts Extraction Pipeline}
\label{appendix:facts_extraction}

Our controlled experiments in \S \ref{sec:analysis} require multiple post-processing steps to the reasoning trace. We use Gemini-2.5-Flash for for this purpose, and since those steps are often not trivial we enable reasoning. For fact verification we also enable search. The only exceptions are 2 steps (\S \ref{appendix:extracting_facts} and \S \ref{appendix:answer_filter}) which we defined as critical and especially challenging which were done with Gemini-2.5-Pro.

\subsubsection{Extracting Facts}
\label{appendix:extracting_facts}

% Setup for the listing to ensure it wraps correctly and looks clean
% \lstset{
%     basicstyle=\ttfamily\small,
%     breaklines=true,
%     frame=single, % Adds a border around the prompt
%     columns=fullflexible,
%     keepspaces=true
% }

\lstset{
    basicstyle=\ttfamily\small,
    breaklines=true,
    breakindent=0pt, % Removes the indent on wrapped lines
    frame=single,
    columns=fullflexible,
    keepspaces=true
}

% \begin{figure}[htbp]
\begin{figure}[!t]
\begin{lstlisting}
I will give you the content of a thinking process. Your goal is to extract only the concrete, self-contained facts from it.

Thinking content: {thinking_content}

Instructions:
A fact is a complete sentence that states a specific piece of information. It is NOT a standalone keyword, topic, name, or entity. It is also not part of an interpretation of the question. It is also not a part of a solution planning, or a step in a reasoning process. Each fact should be self-contained, and its meaning should not rely on any additional context. 

The following are examples of facts:
- The capital of Iceland is Reykjavik
- Marie Curie was the first woman to win a Nobel Prize
- Neil Armstrong was born on August 5, 1930

The following are examples of non-facts:
- The questions asks about the capital of Iceland.
- The user wants to know the author of a paper.
- The date of the event in October 2024.
- Search engines are best for this.
- Marie Curie
- capital cities
- The year is 2001
- Wikipedia
- day

Output a list of facts. Do not add any information that is not mentioned in the thinking content. In case there are no facts in the thinking content, output NONE.
\end{lstlisting}
\caption{Prompt used for extracting concrete, self-contained facts from a thinking process.}
\label{fig:fact_extraction_prompt}
\end{figure}

\Cref{fig:fact_extraction_prompt} presents the fact-extraction prompt. For this step we use Gemini-2.5-Pro rather than Gemini-2.5-Flash since this stage is critical for the entire analysis. 

\subsubsection{Removing Facts That Appear in the Question}
\label{appendix:question_filter}

% Setup for the listing to ensure it wraps correctly and looks clean
% \lstset{
%     basicstyle=\ttfamily\small,
%     breaklines=true,
%     frame=single, % Adds a border around the prompt
%     columns=fullflexible,
%     keepspaces=true
% }

\lstset{
    basicstyle=\ttfamily\small,
    breaklines=true,
    breakindent=0pt, % Removes the indent on wrapped lines
    frame=single,
    columns=fullflexible,
    keepspaces=true
}

% \begin{figure}[htbp]
\begin{figure}[!t]
\begin{lstlisting}
You are a helpful assistant that filters redundant information.

### Inputs
1. Question: {question}
2. Original Facts:
{summary}

### Task
Analyze the "Original Facts" and determine which to keep based on strict redundancy logic:
1. **REMOVE** a fact ONLY if **all** the information it contains is already explicitly stated in the "Question".
2. **KEEP** a fact if it provides **any** new information, details, or context not found in the "Question" (even if it partially repeats the question).

### Output Rules
1. Output only the filtered list of kept facts.
2. Maintain the formatting of the original list (e.g., if input is bullet points, output bullet points).
3. Do not add any additional introductory text or explanations.
4. If no facts remain after filtering, or if the original list was empty (NONE), output exactly: NONE
\end{lstlisting}
\caption{Prompt used for filtering-out facts that restate information that was present in the question.}
\label{fig:question_filter_prompt}
\end{figure}

Since reasoning traces can restate information from the question while we aim to capture facts introduced during reasoning, we remove such restatements from the extracted list with an additional LLM-based step. We use Gemini-2.5-Flash with reasoning enabled and the prompt from \Cref{fig:question_filter_prompt}.

\subsubsection{Removing Facts That Reveal The Answer}
\label{appendix:answer_filter}

% Setup for the listing to ensure it wraps correctly and looks clean
% \lstset{
%     basicstyle=\ttfamily\small,
%     breaklines=true,
%     frame=single, % Adds a border around the prompt
%     columns=fullflexible,
%     keepspaces=true
% }

\lstset{
    basicstyle=\ttfamily\small,
    breaklines=true,
    breakindent=0pt, % Removes the indent on wrapped lines
    frame=single,
    columns=fullflexible,
    keepspaces=true
}

% \begin{figure}[htbp]
\begin{figure}[!t]
\begin{lstlisting}
Below is a Question, a Target Answer and a list of facts.

Question: "{question}"
Target Answer: "{answer}"

Facts list: 
{thinking_content}

### Task
Identify and remove ONLY the facts that explicitly link the Target Answer to the Question. 
You must differentiate between mentioning the answer and resolving the question.

### Rules
1. Remove the fact if it states or implies that "{answer}" is the answer to the question "{question}".
2. Keep the fact if it simply mentions "{answer}" in a context unrelated to the specific question.
3. Keep the fact if it does not mention the answer at all.

### Output
Output the filtered list of facts. Do not output any other text or explanations.
\end{lstlisting}
\caption{Prompt used for filtering-out facts that contain the final answer and link it to the question.}
\label{fig:answer_filter_prompt}
\end{figure}

% Setup for the listing to ensure it wraps correctly and looks clean
% \lstset{
%     basicstyle=\ttfamily\small,
%     breaklines=true,
%     frame=single, % Adds a border around the prompt
%     columns=fullflexible,
%     keepspaces=true
% }

\lstset{
    basicstyle=\ttfamily\small,
    breaklines=true,
    breakindent=0pt, % Removes the indent on wrapped lines
    frame=single,
    columns=fullflexible,
    keepspaces=true
}

% \begin{figure}[htbp]
\begin{figure}[!t]
\begin{lstlisting}
Below is a list of facts.

List of facts:
{thinking_content}

Your task:
Filter-out facts for which one of the following is true:
1. Mention "{answer}".
2. Mention "{answer}" at higher specificity.
3. Mention "{answer}" at lower specificity.

Output the remaining facts as list, do not add any information based on your knowledge. If there are no facts left, output NONE.

Output just the new list of facts, without additional content.
\end{lstlisting}
\caption{Prompt used for filtering-out facts that contain the final answer.}
\label{fig:answer_filter_conservative_prompt}
\end{figure}

The model may commit to an answer during reasoning, explicitly stating its intention to output it. We therefore remove answer-disclosing statements using the prompt from \Cref{fig:answer_filter_prompt} and run it twice: using the gold answer and the model’s predicted answer, which may differ. Since removing every occurrence of the answer would be overly aggressive, as the model may mention it as part of recall rather than as a committed resolution, we instruct the model to remove only statements that explicitly link the target answer to the question. Since this task is relatively challenging, we use Gemini-2.5-Pro with reasoning enabled and found it to perform extremely well. 

 \begin{figure*}[t] 
    \centering
    \includegraphics[width=0.75\textwidth, keepaspectratio]{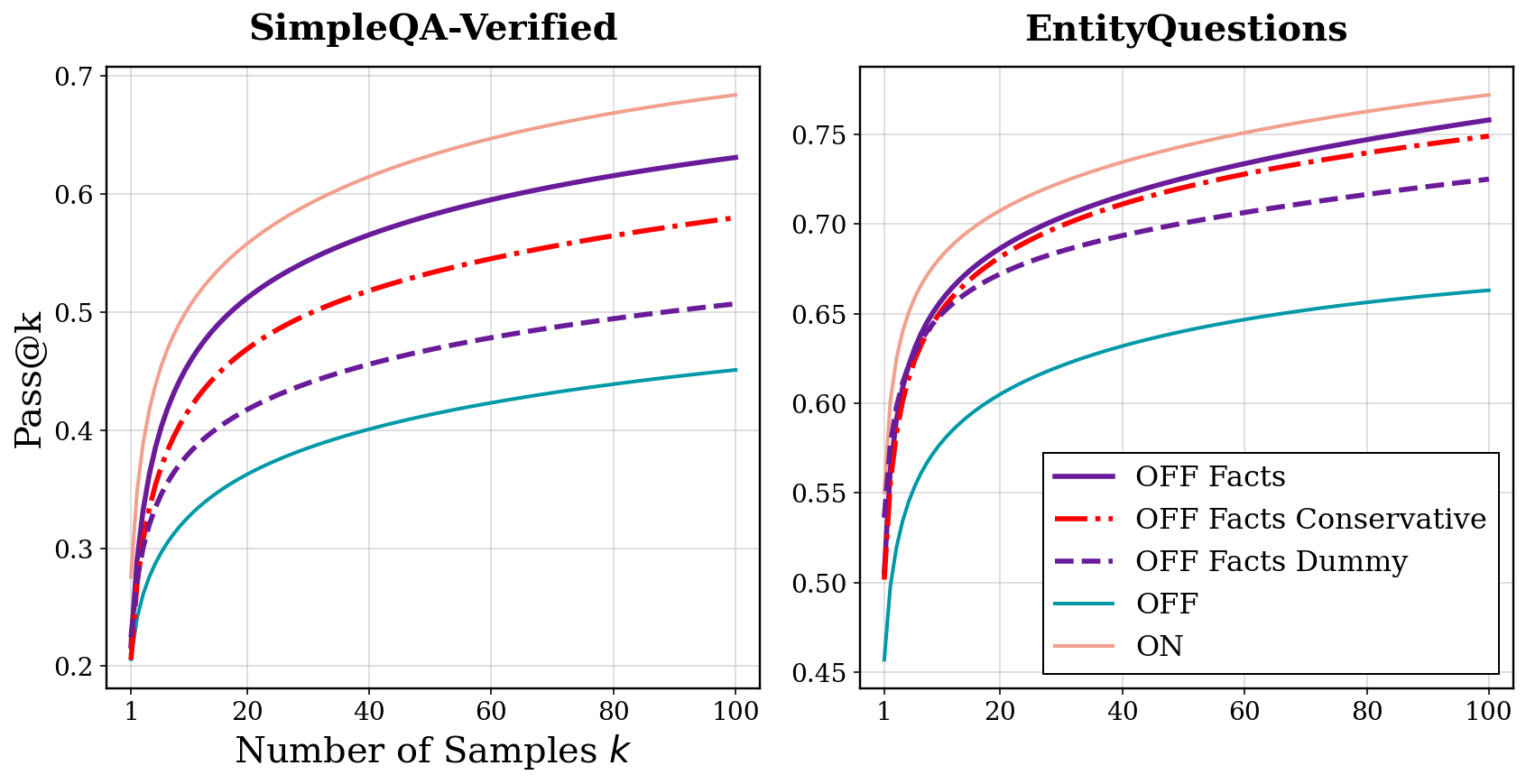}
    \caption{
    Factual priming effect on Gemini-2.5-Flash (\S\ref{sec:factual_priming}), conditioning on facts recalled during reasoning, with reasoning \texttt{OFF}. \texttt{OFF\,Facts} is our main baseline used throughout the paper, while \texttt{OFF\,Facts\,Conservative} is provided as additional reference with \textit{conservative} filtering, where each mention of the answer is removed (see \S \ref{appendix:question_filter}.}
    \label{fig:off_summary_plot_conservative}
    % \vspace{-5pt}
\end{figure*}

As an additional reference, we also checked a \textit{conservative} prompt, presented in \Cref{fig:answer_filter_conservative_prompt}, where the model is instructed to filter out any fact that contains the answer. In \Cref{fig:off_summary_plot_conservative}, we plot the \texttt{OFF\,Facts\,Conservative} baseline in addition to \texttt{OFF\,Facts}. As expected, it underperforms \texttt{OFF\,Facts}, since we filter-out important mentions of the candidate answer. Nevertheless, it still considerably ourperforms the dummy baseline which reinforces our evidence of the effectiveness of the facts themselves.

\subsubsection{Detecting Hallucinated Facts}
\label{appendix:detecting_hallucinations}

% --- Optional: Styling for the code block ---
\lstset{
    basicstyle=\ttfamily\small, % Uses a standard code font
    frame=single,               % Adds a border box around the text
    breaklines=true             % Automatically wraps long lines
}

\begin{figure}[!t]
\begin{lstlisting}
Below is a list of facts, parse them and return a JSON array of strings with each fact.

Ouput format is like this:

```json
[
  "Fact 1.",
  "Fact 2."
]
Facts:
{facts}
\end{lstlisting}
\caption{Prompt used for parsing the list of facts.}
\label{fig:parse_facts_prompt}
\end{figure}

% Setup for the listing to ensure it wraps correctly and looks clean
% \lstset{
%     basicstyle=\ttfamily\small,
%     breaklines=true,
%     frame=single, % Adds a border around the prompt
%     columns=fullflexible,
%     keepspaces=true
% }

\lstset{
    basicstyle=\ttfamily\small,
    breaklines=true,
    breakindent=0pt, % Removes the indent on wrapped lines
    frame=single,
    columns=fullflexible,
    keepspaces=true
}

% \begin{figure}[htbp]
\begin{figure}[!t]
\begin{lstlisting}
Check the correctness of the fact below by looking for supporting evidence using search.

If you find supporting evidence then conclude that it is correct.
If you find contradicting evidence then conclude that it is incorrect.
If you fail to find evidence then conclude that it is incorrect.
If the fact cannot be fully understood without without additional context, then conclude that it is illegal.
If the correctness is inconclusive based on the evidence and the provided context, then conclude that it is unknown.

Fact: {fact}

Finish your response with a new line that contains only your final conclusion ("correct", "incorrect", "illegal" or "unknown") with no additional text around it.
\end{lstlisting}
\caption{Prompt used for detecting hallucinations.}
\label{fig:detect_hallucinations_prompt}
\end{figure}

To verify each fact in the reasoning trace we run Gemini-2.5-Flash with search enabled by using the prompt from \Cref{fig:detect_hallucinations_prompt}. Since our original facts lists are actually strings and not lists, we first  prompt gemini to parse the facts string into json, so we could programatically iterate over them and invoke Gemini on each fact. We manually verified 10 facts that were labeled as correct and 10 facts that were labeled as wrong and found all the labels to be accurate.

% \subsection{Technical Details on the \texttt{OFF\,Facts} and \texttt{ON\,Facts} Baselines}
\subsection{Technical Details on the ``OFF Facts'' and ``ON Facts'' Baselines}
\label{appendix:off_facts}

In the \texttt{OFF\,Facts} baseline we run the model with reasoning disabled but provide the facts list as additional context. The prompt is presented in 
\Cref{fig:off_facts_prompt}, the ``$\{\texttt{context}\}$'' parameter is replaced with the facts list. In case of \texttt{OFF\,Facts\,Dummy} we repeat the string \textit{``There is no factual information in the context.''} until matching the length of the list of facts and provide the resulting string as ``$\{\texttt{context}\}$'' see example in \Cref{fig:off_summary_case_study}).

In the \texttt{ON\,Facts} baseline we simulate a condition as if the model generated a thought that looks like our list of facts. This requires to override the thought with this facts list and rerun the answer prediction step. In case of \texttt{ON\,Facts\,Dummy} we repeat the string \textit{``Let me think.''} until matching the length of the list of facts and override the thought with the resulting string.

 \begin{figure*}[t] 
    \centering
    \includegraphics[width=0.8\textwidth, keepaspectratio]{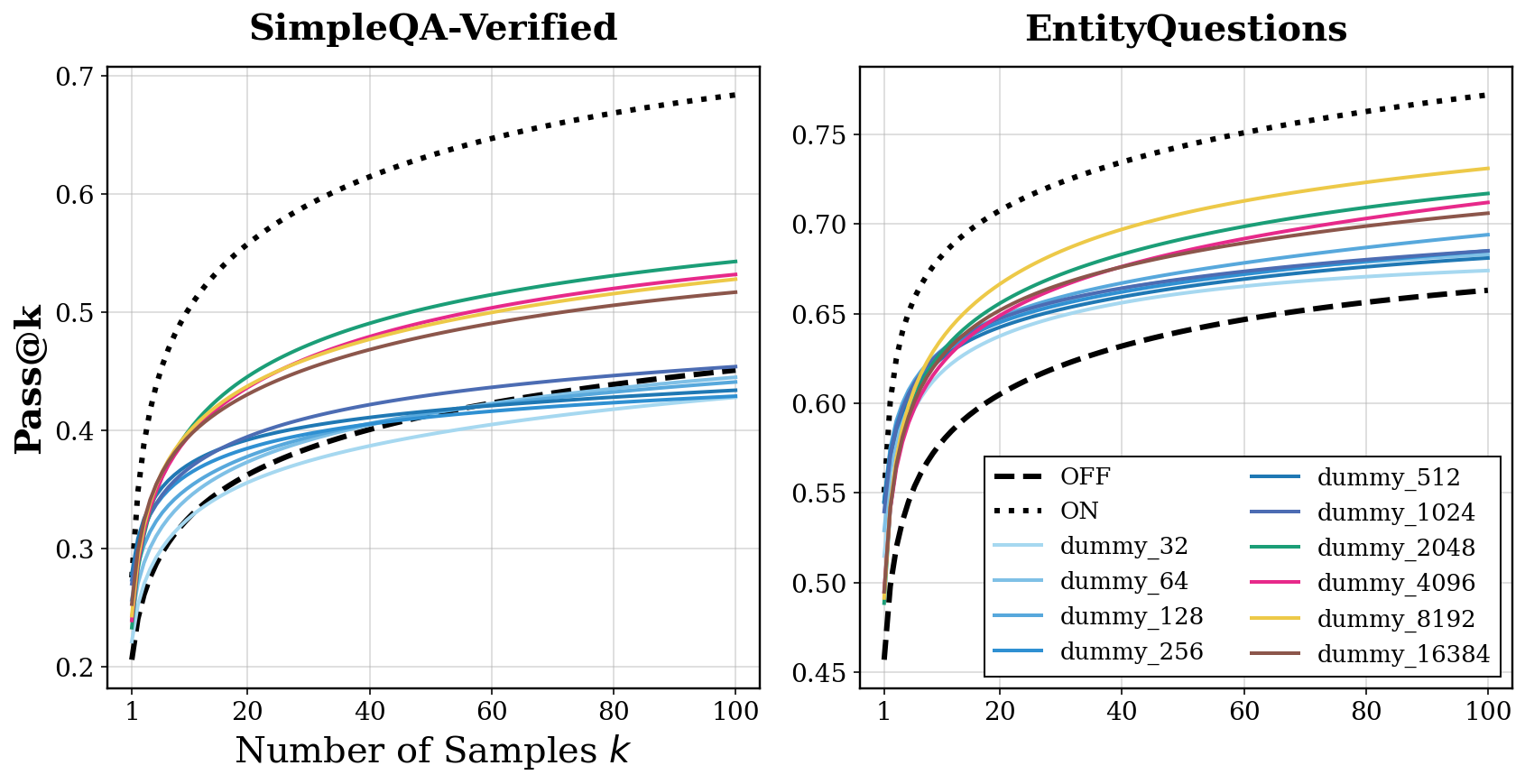}
    \caption{The effect of increasing compute. \texttt{dummy\_X} overrides the thinking trace with a short dummy sequence which is repeated such that the total input length will be X.}
    \label{fig:dummy_scaling}
    \vspace{-5pt}
\end{figure*}

% Setup for the listing to ensure it wraps correctly and looks clean
% \lstset{
%     basicstyle=\ttfamily\small,
%     breaklines=true,
%     frame=single, % Adds a border around the prompt
%     columns=fullflexible,
%     keepspaces=true
% }

\lstset{
    basicstyle=\ttfamily\small,
    breaklines=true,
    breakindent=0pt, % Removes the indent on wrapped lines
    frame=single,
    columns=fullflexible,
    keepspaces=true
}

% \begin{figure}[htbp]
\begin{figure}[!t]
\begin{lstlisting}
I will give you a question, and additional context with factual information.

Question: {question}
Additional context: {context}

You need to predict the answer to the question.
You can use the provided context to help inform your answer if it is relevant. 

If the context is empty, unhelpful, or contains no factual information, ignore it and answer the question using your internal knowledge alone.
\end{lstlisting}
\caption{Prompt used for running the model with reasoning disabled but conditioned on the facts list as additional context.}
\label{fig:off_facts_prompt}
\end{figure}

\end{document}